%% file: main.tex
\documentclass[format=acmsmall, review=false, screen=true]{acmart}

\usepackage{booktabs} 
\usepackage{multirow}
\usepackage{bm}
\usepackage[ruled]{algorithm2e} 

\SetAlFnt{\small}
\SetAlCapFnt{\small}
\SetAlCapNameFnt{\small}
\SetAlCapHSkip{0pt}
\IncMargin{-\parindent}

\acmJournal{TOIS}
\acmVolume{1}
\acmNumber{1}
\acmArticle{11}
\acmYear{2018}
\acmMonth{4}
\copyrightyear{2018}

\setcopyright{acmlicensed}

\acmDOI{0000001.0000001}

\received{February 2018}

\begin{document}
\title[Memory-augmented Dialogue Management for Task-oriented Dialogue Systems]{Memory-augmented Dialogue Management for Task-oriented Dialogue Systems}
\thanks{This work was partly supported by the National Basic Research Program (973 Program) under grant No. 2013CB329403, and the National Science Foundation of China under grant No. 61272227/61332007.}
\author{Zheng Zhang}
\author{Minlie Huang}
\authornote{Corresponding author}
\affiliation{%
  \institution{Tsinghua University}
  \department{Department of Computer Science and Technology}
  \city{Beijing}
  \postcode{100084}}
\email{zhangz.goal@gmail.com}
\email{aihuang@tsinghua.edu.cn}

\author{Zhongzhou Zhao}
\author{Feng Ji}
\author{Haiqing Chen}
\affiliation{%
  \institution{Alibaba Group}
  \city{Hangzhou}
  \state{Zhejiang}
  \postcode{311121}}
\email{zhongzhou.zhaozz@alibaba-inc.com}
\email{zhongxiu.jf@alibaba-inc.com}
\email{haiqing.chenhq@alibaba-inc.com}

\author{Xiaoyan Zhu}
\affiliation{%
  \institution{Tsinghua University}
  \department{Department of Computer Science and Technology}
  \city{Beijing}
  \postcode{100084}}
\email{zxy-dcs@tsinghua.edu.cn}

\begin{abstract}
Dialogue management (DM) decides the next action of a dialogue system according to the current dialogue state, and thus plays a central role in task-oriented dialogue systems. 
Since dialogue management requires to have access to not only local utterances, but also the global semantics of the entire dialogue session, modeling the long-range history information is a critical issue. To this end, we propose a novel Memory-Augmented Dialogue management model (MAD) which employs a memory controller and two additional memory structures, i.e., a slot-value memory and an external memory. The slot-value memory tracks the dialogue state by memorizing and updating the values of semantic slots (for instance, {\it cuisine, price, and location }), and the external memory augments the representation of hidden states of traditional recurrent neural networks through storing more context information. To update the dialogue state efficiently, we also propose slot-level attention on user utterances to extract specific semantic information for each slot.
Experiments show that our model can obtain state-of-the-art performance and outperforms existing baselines.
\end{abstract}

%
%


%
%


\maketitle

\renewcommand{\shortauthors}{Z. Zhang et al.}

\input{samplebody-journals}

\end{document}

%% file: samplebody-journals.tex
\section{Introduction}
Task-oriented dialogue systems offer a natural and effective interface for users to seek information and complete complex tasks in an interactive manner. Such systems often collect users' preferences in the course of dialogue before issuing the final query to the knowledge base (such as booking a flight ticket). There are also some works \cite{hixon2015learning, saha2018complex} viewing the task-oriented dialogue task as a context-aware, multi-turn question answering (QA) task in which a user can interact with the system in multi-turn contexts and the system also has access to the knowledge base.

\begin{figure}
  \includegraphics[width=0.6\textwidth]{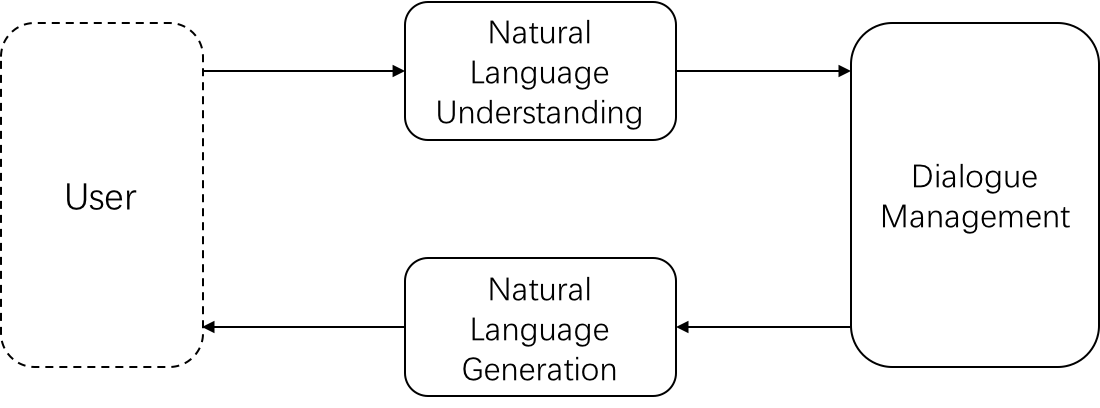}
  \caption{The processing flow of task-oriented a dialogue system. Natural language understanding (NLU) parses the user utterance and extracts structured semantic information from the utterance, dialogue management receives the semantic information and decides the next dialogue act that the system should take, and natural language generation (NLG) translates the dialogue act to a natural language response. In some cases, NLU and DM can be coupled together as a single module, and the semantic information produced by NLG is often unstructured in this situation, such as the output of neural network.
  }
  \label{fig:dialog_system}
\end{figure}

 Different from open-domain conversational systems which are often modeled in an end-to-end manner, task-oriented dialogue systems are generally composed of several cascaded processes, as shown in Figure \ref{fig:dialog_system}, including natural language understanding (NLU), dialogue management (DM), and natural language generation (NLG).
Dialogue management, which is in charge of selecting actions in response to user inputs, plays a central role in task-oriented dialogue systems \cite{williams2007partially, ge2dialogue}. It takes as input the user intent which is analyzed by NLU, interacts with knowledge base, and decides the next system action. Sometimes NLU and DM can be coupled together as a single module which can be trained end-to-end to read directly from user utterance and produce system action. The system action produced by DM will be translated into a natural language utterance by NLG \cite{wen2015semantically} to interact with users. 

In order to decide the next action a dialogue system should take, dialogue management, particularly in task-oriented dialogue systems, should deal with the dialogue context information. It needs to access not only local utterances, but also the global information about what has been addressed several turns ago. The global history information, which is often referred to as \emph{dialogue state}, is a key factor in dialogue systems. Based on the dialogue state, the dialogue manager then produces system action according to its policy. The task of dialogue management is sometimes divided into two subtasks, namely {\it dialogue state tracking} which maintains dialogue history information, and {\it dialogue policy} which selects the next system action based on the dialogue state. 

Early methods of modeling dialogue management are mostly rule-based, in which the state update and dialogue policy process are manually defined, but these methods did not take into account the probability uncertainty in dialogue. Bayesian network methods \cite{paek2005markov, williams2007partially} formulated dialogue management as a probabilistic graphical model which models the conditional dependency between different states, and each specific state is bound with an action to be taken, but the definition of dialogue state still need manually-crafted rules. Recently, many neural network methods have been proposed for dialogue management due to their capability of semantic representation and automatic feature extraction, and obtain state-of-the-art performance on many dialogue tasks \cite{ge2015dialogue, serban2016building}. More specifically, most neural dialogue models are RNN (Recurrent Neural Network) based which takes as input user utterance and system response at each dialogue turn, and the hidden state of RNN is utilized as the representation of dialogue state \cite{henderson2014word, williams2017hybrid}.

However, despite of the success of RNN on various text modeling tasks, simple RNN is proven to have poor performance on dialogue tasks \cite{williams2017hybrid} due to the single hidden state vector used in RNN and thus the defect of modeling long-range contexts. Hierarchical RNN structures ~\cite{serban2016hierarchical} and memory networks~\cite{weston2014memory,dodge2015evaluating, bordes2016learning} are feasible solutions to this issue, but existing neural models still lack an explicit memorization of the history semantics of the entire dialogue session: the dialogue act types, semantic slots, and the values of the slots are not explicitly processed during the interaction.

\begin{figure}
  \includegraphics[width=0.5\textwidth]{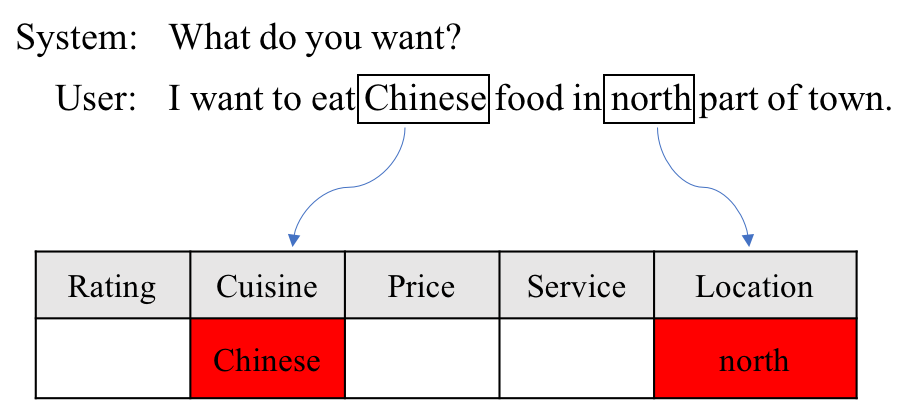}
  \caption{Slot-level attention: word mentions in user utterance are mapped to semantic slots such as {\it rating, cuisine, price, service,} and {\it location}.
  }
  \label{fig:word_attn}
\end{figure}

Another important issue is to extract semantic information from user utterance when combining NLU and DM together, which is the case in most end-to-end dialogue systems. Such semantic information is critical for dialogue state update. Existing methods either extract information from predefined features (such as POS and NER tags) by heuristic rules \cite{henderson2014word}, or from pretrained word embeddings by neural network encoder \cite{mrkvsic2016neural}.
However, words in user utterance have different importance for updating dialogue states and predicting the next action, which is not taken into consideration by previous methods. For example, in a user utterance \emph{I want to book a table in Beijing Hotel}, the word \emph{book} apparently contributes more than the word \emph{want} to the user intent. Furthermore, each word contributes differently to different slots, e.g., word \emph{British} is more related to slot \emph{Cuisine} while \emph{north} is more related to \emph{Location}, as shown in Figure \ref{fig:word_attn}.


To address the above issues, we propose a novel Memory-Augmented Dialogue management model (MAD) which attentively receives user utterances as input and predicts the next dialogue act\footnote{The dialogue act can be translated into a natural language utterance by a language generator, as shown in \cite{wen2015semantically}.
}. 
Dialogue act is composed of two parts in our model: {\it dialogue act type} and {\it slot-value pairs}, as shown in Table \ref{tab:dialogact} . 
Dialogue act type indicates the intent type such as {\it Query} or {\it Recommendation}, which is a high-level representation of dialogue act. Slot-value pairs denote key elements of a task, and represent the key semantic information supplied by the user during the interaction, which also indicate the state of the dialogue.

We design two memory modules, namely a slot-value memory and an external memory, which can be written (or updated) and read, to enhance the ability of modeling history semantics of dialogues. 
A memory controller is introduced to control the write and read operations to the two memories.
The slot-value memory explicitly memorizes and updates the values of the semantic slots during interaction. The write to the slot-value memory units, each corresponding to a slot, is implemented by a slot-level attention mechanism. In this manner, the slot-value memory provides an observable and interpretable representation of the dialogue state. The external memory serves as a supplement to the single hidden state of a RNN structure and provides a better capacity to store more historical dialogue information. A complete dialogue act (consisting of dialogue act type and slot-value pairs) for the next interaction is predicted based on the slot-value memory and external memory.

\begin{table}[htbp]
\small
\centering
\begin{tabular}{|c|c|c|c|c|c|}
  \hline
    {\bf Utterance} & \multicolumn{5}{c|}{How about a {\em British} restaurant in {\em north} part of town.}\\
  \hline
    {\bf Dialogue act type}&\multicolumn{5}{c|}{\em Query}\\
  \hline
    {\bf Slot-value pairs}&\multicolumn{5}{c|}{Cuisine={\em British}, Location={\em Paris}}\\
  \hline
    \multirow{2}{*}{\bf Mask} (auxiliary) & Rating & Cuisine & Price & Service & Location\\
  \cline{2-6}
    & 0 & 1 & 0 & 0 & 1\\
  \hline
\end{tabular}
\caption{\label{tab:dialogact}An example of dialogue act for a given utterance. Dialogue act type is a high-level representation of an utterance. Slot-value pairs are the task-specific semantic elements that are mentioned in an utterance.}
\label{appendix-da-example}
\end{table}



Our contributions are summarized as follows:
\begin{itemize}
    \item We propose a novel memory-augmented dialogue management model by introducing two memory networks. The slot-value memory network maintains the values of semantic slots during interaction, and the external-memory augments the single state representation of the recurrent networks. Both memory modules enable the model to access not only local utterances, but also the global semantics of the entire dialogue session. 
    
    \item We propose an attention mechanism for updating the dialogue state. In particular, the model first computes a weight distribution over all words in a user utterance for each slot. Then, the weighted representation of the utterance is used to update the memory unit for each slot.
    
    \item The model can offer more observable and interpretable results in that the slot-value memory can track the change of dialogue states explicitly.
    
\end{itemize}

\section{Related Work}
\label{sec:rel-work}

The role of dialogue management (DM) is to launch the next interaction through predicting the next action the system should take, or by generating an utterance directly in response to user's query. The previous studies on DM can be broadly classified into three types: rule-based models, Bayesian network models, and neural models.

Rule-based approaches date back to very early dialogue systems~\cite{weizenbaum1966eliza}. Several architectures are proposed to formulate the process of dialogue management. The \emph{flow diagram} approach \cite{mctear1998modelling} used a finite-state machine to model state transition in dialogue, where the state represents a certain dialogue status, and the transition between states is triggered by the corresponding type of a user utterance.  \emph{Slot-filling} approaches \cite{goddeau1996form} expanded the definition of dialogue state to an aggregation of slots and values. In such models, user can talk about each slot by issuing constraints and requesting the values of slots, and the dialogue state will be updated as long as a user provides new values for the slots during interaction.
Though \emph{rule-based} DM models work well in some applications, these approaches have apparent difficulties in task and domain adaptation~\cite{zukerman2001predictive} because the rules are usually tailored to a specific scenario. Due to the nature of hand-crafted rules, the variety and diversity of language is not well addressed. The need for hand-crafted rules also makes it expensive to build a {\em rule-based} system. 

Bayesian network approaches are proposed to address the issues of rule-based methods. Dialogue management was firstly formalized as a Markov decision process (MDP) \cite{levin1998using} under the Markov assumption \cite{paek2005markov}, in which the new state $s_t$ at turn $t$ is only conditioned on the previous state $s_{t-1}$ and system action $a_{t-1}$. MDP models the uncertainty in dialogue and becomes more robust to the errors induced by speech recognition and NLU. Partially observable Markov decision processes (POMDP) \cite{williams2007partially} provides a more principled way in that it takes environment observation $o_t$ into consideration. On the top of this framework, state transition and dialogue policy are trained using reinforcement learning. However, the POMDP model becomes difficult to train for the domains with large state space. An improved version of POMDP - Hidden Information State (HIS)  \cite{young2007hidden} is proposed to address this problem by grouping dialogue states into partitions. Another key problem in building Bayesian dialogue model is the lack of training corpus, thus user simulation \cite{schatzmann2006survey} is employed to enhance the training procedure, where dialogue data can be collected through interactions between a user simulator and a target system.
In spite of the success of Bayesian network methods, designing an appropriate reward function and manually crafting features limit the applicability of such approaches. As a noticeable defect, the state in these approaches is still manually defined, requiring a large amount of human labor.

A variety of neural models have recently been applied for the dialogue management task. Since the process of a dialogue session naturally follows a sequence-to-sequence learning problem at the turn level, recurrent neural network (RNN) is proposed to model the process \cite{henderson2014word, mrkvsic2016neural, wen2017latent}. At each turn, RNN takes as input the structured semantic representation produced by NLU (or raw user utterance when combining NLU and DM together) and predicts system action, where the hidden state of RNN is utilized as the representation of a dialogue state. 
There are also some neural end-to-end models which directly take dialogue context as input and generate natural language response \cite{shang2015neural, li2016deep, serban2016building, serban2017hierarchical} in open-domain conversational systems.
However, due to the vanishing gradient problem and the limited ability of state representation, RNN is difficult to capture the long-range context in dialogue. Hybrid Code Networks \cite{williams2017hybrid} proposes to handle the state representation problem by combining rule-based and RNN-based models together, while the performance is still highly dependent on the hand-crafted rules.



Memory network provides a principled approach for modeling long-range dependency and making multi-hop reasoning, which has advanced many NLP tasks such as machine translation \cite{wang2016memory} and question answering \cite{sukhbaatar2015end}. Neural turing machines \cite{graves2014neural} was proposed to augment existing neural models with additional memory units to solve complicated tasks. It is analogous to a Turing machine but is differentiable end-to-end. \cite{weston2014memory} proposed \emph{fully supervised memory networks} which employ supervision signal not only from answer labels but also from pre-specified supporting facts. \cite{sukhbaatar2015end} proposed \emph{end-to-end memory networks} (MEMN2N) which can be trained end-to-end without any intervention on which supporting fact should be used during training. \emph{Dynamic memory network} proposed by \cite{kumar2016ask} uses a sentence-level attention mechanism to update its internal memory during multi-hop inference. \emph{Key-value memory network} \cite{miller2016key} encodes prior knowledge by introducing a key memory structure which stores facts to address to the relevant memory value. There are already some works which introduced memory network into the task of dialogue management \cite{perez2016dialog} where memory networks are straightforwardly applied in a machine reading manner. In comparison, our model is better to model the long-range history semantics of the dialogue session by memorizing and updating the dialogue act types and the values of semantic slots explicitly, which is implemented through a slot-value memory and an external memory.

Extracting semantic information from user utterance is a key issue in task-oriented dialogue systems when combining NLU and DM together.
Early methods used hand-crafted rules and semantic features, including NER and POS tags, to construct semantic features for user utterance. \cite{henderson2014word} proposed to use the speech recognition confidence score as an additional feature. \cite{serban2016building, serban2017hierarchical} used hierarchical RNN models, where the user utterance is processed by a word-level RNN, and utterances are sequentially connected through an utterance-level RNN. \cite{mrkvsic2016neural} proposed to use convolutional neural network (CNN) model for semantic feature extraction. However, existing approaches did not consider the fact that words in an utterance contribute differently to different slots,  which is important for updating the dialogue state.


\section{Memory-augmented Dialogue Management with Slot-Attention}

\begin{figure}
  \includegraphics[width=0.6\textwidth]{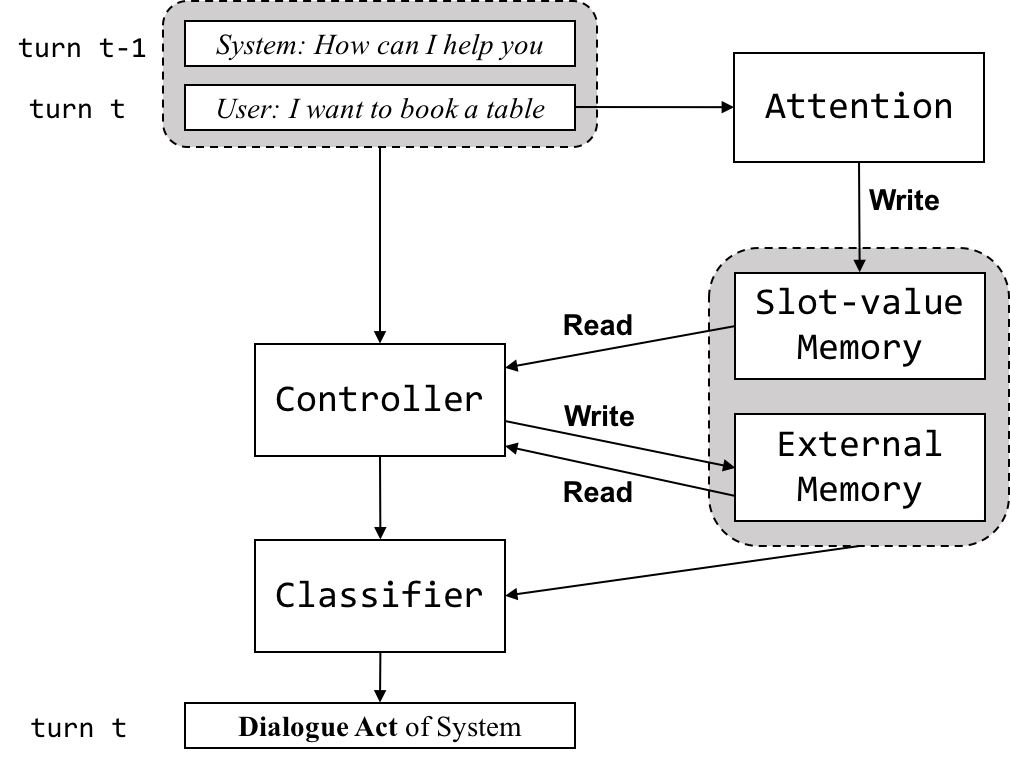}
  \caption{Memory-augmented Dialogue Management (MAD): At each dialogue turn $t$, the model takes as input the current user utterance and the previous system response, and predicts the next dialogue act. The slot-value memory is updated with an attentive read of the user utterance by a slot-level attention mechanism while the external memory is read and updated by the controller. The memory controller along with the two memory modules will predict the next dialogue act of the system by a classifier.
  }
  \label{fig:model}
\end{figure}

\subsection{Task Definition}
\label{sec:taskdef}

This paper deals with task-oriented dialogue management. We start by defining the input and output of our model. At the current turn ($t$) of a dialogue, given a user utterance along with the system response of the previous turn ($t-1$), the task of dialogue management module is to predict the next system dialogue act that will be utilized to generate a natural language utterance. This procedure can be formalized as follows:
\begin{equation*}
    P_{\theta}({DA}_t|x_1, y_1,..., x_{t-1}, y_{t-1}, x_{t})
\end{equation*}
where $x_t$ and $y_{t-1}$ are the user utterance at the current turn and system response at the previous turn, respectively, and ${DA}_t$ is the next dialogue act which can be used to generate system response. $\theta$ represents the parameters of the model. The next system response $y_t$ will be generated from $DA_t$ by a natural language generator, which is beyond the scope of this paper.

To exemplify the concept of {\it dialogue act} in our model, we take the task of restaurant reservation as an example, as shown in Table \ref{tab:dialogact}. Dialogue act ($DA$) is composed of two elements: {\em dialogue act type} and {\em slot-value pairs}. Dialogue act type is a general description of user intents, such as {\em Query} where the user may search for some information, and {\em Recommend} where the user may ask for some recommendations. A slot-value pair represents a filled value for a slot
\footnote{Generally speaking, a slot in task-oriented dialogue systems is a category of semantic features, which defines some key attribute or element for accomplishing a task. } 
, such as Location={\em north}, Price={\em expensive} and Cuisine={\em British}. The slot-value pairs are usually regarded as the state representation in many dialogue state tracking studies \cite{henderson2014word}. During the interaction, the filled value for each slot may be provided or updated by the user, and correspondingly, the dialogue state changes. For instance, when the user says {\em How about a British restaurant in north part of town.}, two slot-value pairs, Cuisine={\em British} and Location={\em north}, will be updated. 
However, not all slot-value pairs which are mentioned in the context are to be addressed in the dialogue act of system response. We thus introduce an auxiliary variable {\em Mask}, which is a one-hot vector with dimension $n^s$ which is the number of slots, to decide which slot-value pairs are to be included in the next dialogue act. 
As shown in Table \ref{tab:dialogact}, the slots appeared in dialogue act are only Cuisine and Location, and their mask value is set to 1. In previous dialogue turns, the value of other slots may have already been mentioned, but their value is useless for the system response of this turn, and their Mask value is 0.
Generally speaking, a dialogue act can be viewed as the structured semantic representation of a natural language sentence.





\subsection{Overview}

As shown in Figure \ref{fig:model}, the memory-augmented dialogue management model has two novel memory components, namely slot-value memory $(M^S,M_t^V)$ and external memory $M_t^E$. The slot-value memory consists of a static slot memory ($M^S$) and a dynamic value memory ($M_t^V$) where one memory unit $M^S$(i) in $M^S$ is mapped to a unique unit $M^V$(i) in $M_t^V$. $M^S$ remains unchanged during the interaction, while $M_t^V$ and $M_t^E$ is updated at each turn $t$.
We also design an RNN-based memory controller which controls read and write of the slot-value memory and external memory. 
The slot-value memory is updated with an attentive read of the user utterance by a slot-level attention mechanism while the external memory is read and updated by the controller. The memory controller along with the two memory modules will predict the next dialogue act of the system by a set of classifiers.




Let $x_t=(\mathbf{e}_1^x,...,\mathbf{e}_{n_{x,t}}^x)$ and $y_{t-1}=(\mathbf{e}_1^y,...,\mathbf{e}_{n_{y,t-1}}^y)$\footnote{Note that $y_{t-1}$ is the system response at turn $t-1$ while $y_t$ is to be generated with a predicted $DA_t$.} denote the word embedding sequence of the user utterance at turn $t$ and the preceding system response at turn $t-1$, respectively, where $\mathbf{e}_i^x,\mathbf{e}_j^y\in\mathbb{R}^m$ are word embeddings, $n_{x,t}$ and $n_{y,t-1}$ are the lengths of two sequences. At each turn $t$, our model works in the following procedure:
\\
\\{\bf 1. Memory Read}:
The controller reads information from the value memory and external memory. The read of $M_t^V$ is conditioned on the controller state ($S_{t-1}$) and the value memory ($M_{t-1}^V$) at the previous turn, and the slot memory, formally as follows:
\begin{equation}
    \mathbf{r}_t^{V}= \mathbf{read^v}(S_{t-1}, M^S, M_{t-1}^V),\\
\end{equation}
and the read of the external memory conditions on the controller state and the external memory at the previous turn:
\begin{equation}
    \mathbf{r}_t^E= \mathbf{read^e}(S_{t-1}, M_{t-1}^E).
\end{equation}
Inspired by \cite{graves2014neural}, we introduce content-based addressing for memory read. $\mathbf{r}_t^{V},\mathbf{r}_t^E\in\mathbb{R}^m$ are content vectors read from the slot-value memory and the external memory, respectively.
\\
\\{\bf 2. Controller State Update}:
The controller state $S_{t-1}$ is then updated by the information read from the value memory and the external memory, and the content from $x_t$ and $y_{t-1}$:
\begin{equation}
    \mathbf{S}_t=\mathbf{GRU}(\mathbf{S}_{t-1}, [x_t;y_{t-1}; \mathbf{r}_t^V; \mathbf{r}_t^E])
\end{equation}
where GRU stands for gated recurrent units \cite{cho2014learning}, and $[\cdot;\cdot]$ denotes the concatenation of vectors. 
For simplicity, an utterance ($x_t/y_{t-1}$) is represented by the averaged word embeddings but more elaborated representation models are also applicable.
\\
\\{\bf 3. Memory Write}:
Memory vectors in $M_{t}^V$ and $M_{t}^E$ are updated based on $S_t$ and their previous values:
\begin{gather}
    M_t^V=\mathbf{write^v}(S_t, M^S, M_{t-1}^V)\\
    M_t^E=\mathbf{write^e}(S_t, M_{t-1}^E)
\end{gather}
The output at turn $t$ is obtained  based on $S_t$ and $M_t^V$. The output consists of the elements of a dialogue act, that is, the dialogue act type, slot-value pairs and a mask. Note that the slot memory $M^S$ is static and does not need to be updated.


\subsection{Slot-Value Memory}
\label{sec:sv-memory}
The slot-value memory tracks the dialogue state by storing and updating the value of each semantic slot during interaction. It is composed of two components: slot memory and value memory, and both of them are composed of the same number ($n_s$) of column vectors. The slot memory is kept constant during the dialogue, with each column vector $M^S(i)$ corresponding to a semantic slot $i$. The semantic slots are like {\it Location, Price, or Cuisine}. 
Inspired by \cite{miller2016key}, each slot memory unit $M^S$(i) in our model acts as the index, which helps to locate the content in $M_t^V$. In our proposed model, we further apply the slot memory unit to extracting slot-relevant information from user utterance.
Thus we keep $M^S$ unchanged during training and test time, and $M^S(i)$ is initialized by the averaged embeddings of words in slot $i$.

The value memory stores the value of each slot $i$ in $M_t^V(i)$. During the dialogue, the value of a slot may be added into the memory when a new slot is mentioned, or an old value can be updated to a new value of a previously mentioned slot. That is, each memory unit in the value memory stores the latest value (may be empty) of a semantic slot.
\\ \\
{\bf Read from the slot-value memory}
In our model, the main function of the slot-value memory is to trace the latest value of each slot, which is critical for predicting the slot-value pairs in the dialogue act. However, the effect of the slot-value memory on the state update of the controller is not straightforward. Thus, we employ a simple method for the read from the slot-value memory, which is the average of the vectors in the value memory:
\begin{equation}
    \mathbf{r}_t^V=\frac{1}{n_s}\sum_i{M_{t-1}^V(i)},
\end{equation}
where $n_s$ is the number of slots.
\\ \\
{\bf Write to the slot-value memory}
The write to $M_t^V$(i) depends on slot addressing which decides how much information should be updated for each slot when giving a user utterance. Ideally, the value memory is supposed to update its values for all slots that are mentioned in a user utterance. 
For example, when user inputs an utterance "{\em I want a Chinese restaurant}",  the model updates slot {\it Cuisine} with a new value {\em Chinese}.

Inspired by \cite{graves2014neural,miller2016key}, we apply a slot addressing technique to decide the amount of information that should be updated to each value memory vector of the corresponding slot given a user utterance:
\begin{equation}
    M_t^V(i)={\beta_t^i}{\mathbf{c}_t^i}+(1-\beta_t^i)M_{t-1}^V(i)
\end{equation} 
The first term is new information obtained from the attentive representation ($\mathbf{c}_t^i$) of utterance $x_t$ and the second term is the old information maintained. The attentive representation $\mathbf{c}_t^i$ of utterance $x_t$, described soon later, essentially decides the relatedness of the user utterance to slot $i$.
$\beta_t^i$ is a gate which controls how much $M_t^V$ should be updated, and it depends on the attentive read $\mathbf{c}_t^i$ and the last system response $\mathbf{y}_{t-1}$:
\begin{equation} \label{eq:updategate}
    \beta_t^i= {\rm sigmoid}(W_i^c([\mathbf{y}_{t-1};\mathbf{c}_t^i])+b_i^c)
\end{equation}
If utterance $x_t$ mentions slot $i$, $\beta_t^i$ will be large, and the corresponding value memory unit $M_t^V(i)$ will be updated substantially, otherwise much less information will be updated with a smaller $\beta_t^i$. In order to better train these $\beta_t^i$, we employ additional supervision on the weight, as defined in $\mathcal{L}^{att}$ (see Eq. \ref{eq:loss-att}).

\begin{figure}
  \includegraphics[width=0.7\textwidth]{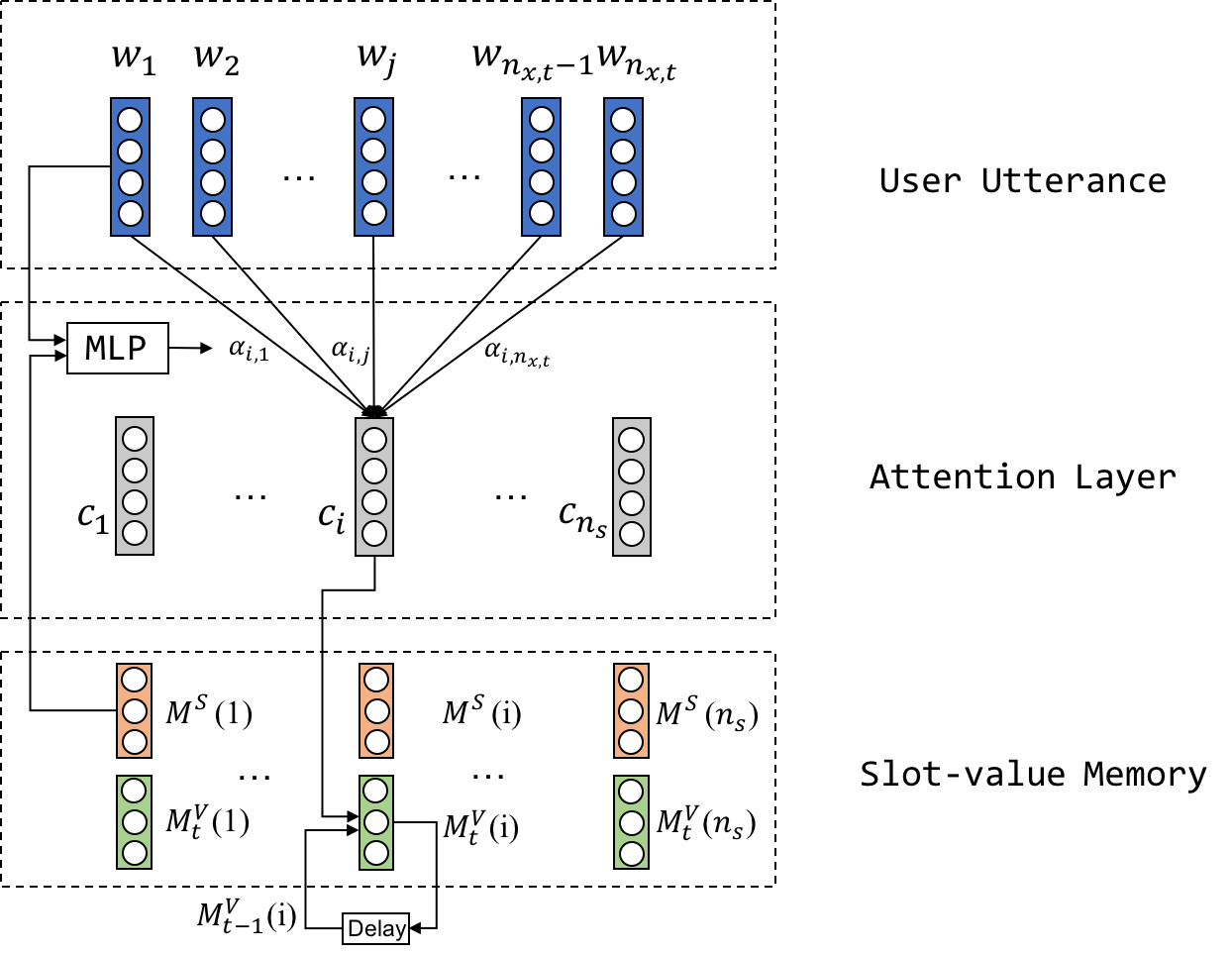}
  \caption{
  Slot-level attention mechanism for updating the slot-value memory.For each slot $i$, the attention score $\alpha_{i,j}$ for each word $j$ is calculated based on word embeddings $e_j$ and slot memory $M^S(i)$. Context vector $c_i$ is the weighted sum of word embeddings of the utterance. Finally, the value memory is updated based on the previous value vector and the context vector. Note that the attention mechanism is applied on each slot $i$.
  }
  \label{fig:slot_attn}
\end{figure}

\subsection{Slot-level Attention}
The context vector $\mathbf{c}_t^i$ in the above section is an attentive representation of utterance $x_t$, conditioned on the $i$-th slot vector. 
More formally, for an user utterance $x_t=(\mathbf{e}_1^x,..,\mathbf{e}_{n_{x,t}}^x)$, we compute attention weights ($\alpha_{i,1},...\alpha_{i,j},...,\alpha_{i,{n_{x,t}}}$) where each weight indicates the similarity of a word embedding $e^x_j$ to a slot memory unit $M^S(i)$, as follows:

\begin{equation}
    \mathbf{c}_t^i=\sum_{j=1}^{n_{x,t}}\alpha_{i,j}\mathbf{e}^x_j \\
\end{equation}
\begin{equation}\label{eq:attnweight}
    \alpha_{i,j}=\frac{\exp({d}_{i,j})}{\sum_{k=1}^{n_{x,t}}\exp({d}_{i,k})} \\
\end{equation}
\begin{equation}
    {d}_{i,j}=MLP([M^S(i), \mathbf{e}^x_j])
\end{equation}
For the previous example, the weight between word {\em Chinese} and slot {\em Cuisine} will be large, while the weights between other words and this slot will be much smaller. The learning of $\alpha_{i,j}$ is also supervised as shown in $\mathcal{L}^{att}$ (see Eq. \ref{eq:loss-att}).

\subsection{External Memory}

The external memory is used to augment the representation capacity of the single state of RNN \city{henderson2014word}, and it is sometimes referred to as {\em memory state} \cite{wang2016memory} in other works. Varies from the slot-value memory, external memory is not endowed with explicit semantic meaning in our framework. The external memory $M_t^E$ consists of $n_e$ columns of $m$-dimensional unit vectors, which are to be read and written to during dialogue controlled by the memory controller.
\\{\bf Read}
The read vector $\mathbf{r}_t^E$ at turn $t$ is a weighted sum of the memory units:
\begin{equation}
    \mathbf{r}_t^E=\sum_{i=1}^{n_e}\mathbf{w}_t^r(i){\cdot}M_{t-1}^E(i)
\end{equation}
where $n_e$ is the number of external memory units. And the weight $\mathbf{w}_t^r\in\mathbb{R}^{n_e}$ is given by
\begin{equation}
\label{eq-read-weight-ext}
    \mathbf{w}_t^r={\bm{g}_t^r}{\cdot}{\mathbf{w}_{t-1}^r}+(\mathbf{1}-\bm{g}_t^r){\cdot}{\widetilde{\mathbf{w}}_t^r}
\end{equation}
where $\mathbf{g}_t^r{\in}\mathbb{R}^{n_e}$ is an update gate which controls the amount of $\mathbf{w}_{t-1}^r$ to be updated, and $\widetilde{\mathbf{w}}_t^r$ is a weight controlling new information to read from $M_{t-1}^E$ conditioned on the state of the controller $\mathbf{S}_{t-1}$.
\begin{gather}
    \mathbf{g}_t^r=\sigma({W_g^r}{\mathbf{S}_{t-1}}) \\
    \widetilde{\mathbf{w}}_t^r={\rm softmax}(\mathbf{v}^\top[M_{t-1}^E(i);\mathbf{S}_{t-1}]) 
\end{gather}
{\bf Write}
There are two operations during the write to the external memory: $erase$ and $add$. $erase$ controls how much old information should be removed from the memory and $add$ controls the addition of new information. Formally, 
\begin{equation}
    M_t^E(i)=M_{t-1}^E(i)(\mathbf{1}-\theta(i){\cdot}\bm{\mu}_t^e)+\theta(i){\cdot}\bm{\mu}_t^a
\end{equation}
where the first term is the left information after erased by vector $\bm{\mu}_t^e \in \mathbb{R}^m$, and the second is new information added by vector $\bm{\mu}_t^a \in \mathbb{R}^m$. The scalar $\theta(i)=\bm{w}_t^r(i)$, the read weight on memory unit $i$, as defined in Eq. \ref{eq-read-weight-ext}.


Both {\it erase vector} and {\it add vector} are obtained conditioned on the state of the controller $S_t$, as follows:
\begin{gather}
    \bm{\mu}_t^e=\sigma(W^e\bm{S}_t)\\
    \bm{\mu}_t^a=\sigma(W^a\bm{S}_t)
\end{gather}

\begin{figure}
  \includegraphics[width=0.7\textwidth]{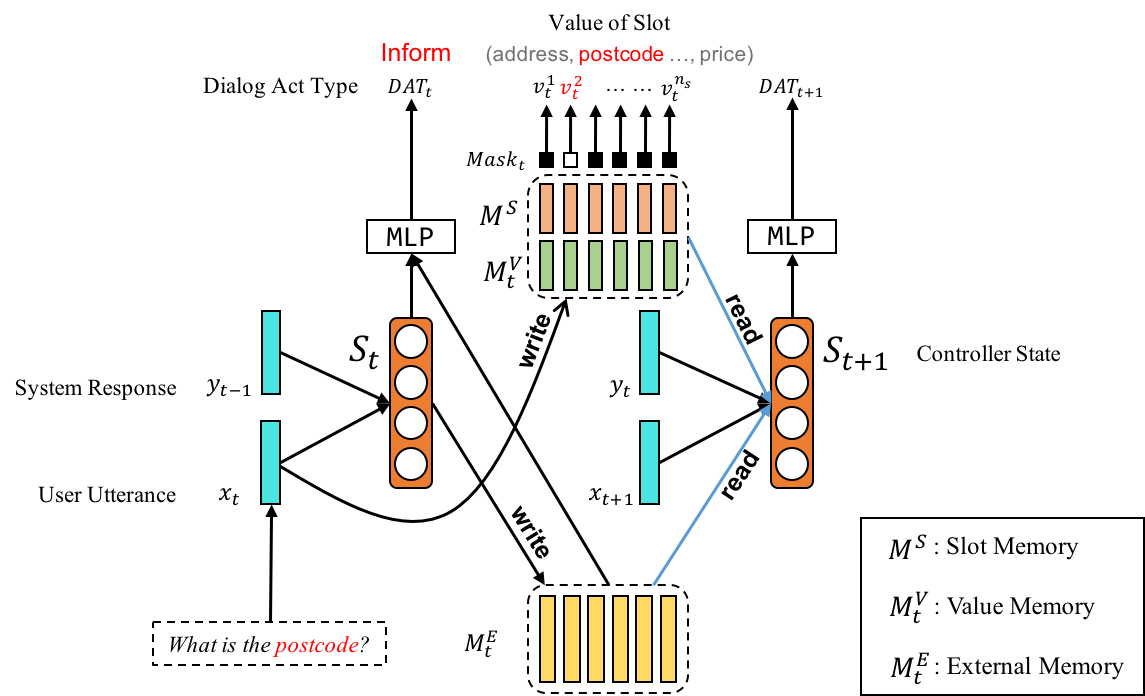}
  \caption{Dialogue act prediction of MAD: ${DAT}_t$ is the dialogue act type of system response at turn $t$. ${Mask}_t$ is the mask for slot-value pairs at turn $t$, and the color of each mask block indicates its value, with white indicating 1 and black for 0. $v_t^i$ represents the value of slot $i$. The prediction of ${Mask}_t^i$ and $v_t^i$ are both based on $M_t^E$(i).
  }
  \label{fig:model2}
\end{figure}

\subsection{Dialogue Act Prediction}
\label{sec:prediction}

As illustrated in Figure. \ref{fig:model2}, our memory-augmented network predicts a dialogue act as follows: first, the dialogue act type is predicted via $P_t^{dat}$; second, each slot is associated with a binary classifier ($P_t^{m,i}$) that decides whether the $i-$th slot should be included in the final dialogue act; third, if a slot $i$ is selected, the value of the slot is predicted by $P_t^i$. The final dialogue act can be assembled by these predicted results. 
%
%
\\
    \\ {\bf Predicting dialogue act type}: this classifier outputs a distribution over dialogue act types such as {\em Inform}, {\em Request}, and {\em Recommendation}.
    It is implemented by a MLP conditioned on the controller state and all memory units:
    \begin{align}
    \label{predict:dat}
        P_t^{dat}(dat|\bm{S}_t,M_t^E, M_t^V)=MLP([\bm{S}_t;M_t^E(1);...;M_t^E(n_e);M_t^V(1);...;M_t^V(n_s)])
    \end{align}
    where $dat$ is one of all the dialogue act types.
%
%
    \\ {\bf Predicting a slot}: there is a slot mask which controls the slots to be included in the final dialogue act.
    There is a binary classifier for each slot $i$ conditioned on the controller state $S_t$, external memory $M_t^E$ and its corresponding value memory unit $M_t^V(i)$):
    \begin{equation}
    \label{predict:mask}
        P_t^{m,i}(z|\bm{S}_t)=MLP([M_t^E(1);...M_t^E(n_e)])
    \end{equation}
    where $z \in \{0,1\}$, $z=1$ indicates that slot $i$ should be included in the next dialogue act. 
    \\ {\bf Predicting the value of a slot}: once we obtain which slot should be included in the dialogue act, we need to decide which value of the slot should be mentioned. This is given by the classifier which estimates a probability distribution over all the values for a slot:
    \begin{equation}
    \label{predict:sv}
        P_t^{i}(v_j^{i}|\bm{S}_t)=MLP([M^S(i), M_t^V(i)])
    \end{equation}
    where $v_j^{i}$ is all the values of slot $i$. 
%
%



\subsection{Loss Function}
We adopt cross entropy as our objective function.
There are three terms in the function corresponding to the prediction of dialogue act types ($\mathcal{L}^{dat}$), slot-value pairs ($\mathcal{L}^{v}$), and slot mask ($\mathcal{L}^{m}$), as presented in the previous section.

The loss function is defined as follows:
\begin{equation}
\label{eq:loss}
    \mathcal{L}=\mathcal{L}^{dat}+\gamma \sum_i\mathcal{L}^m(i)+\lambda\sum_i\mathcal{L}^v(i)
\end{equation}
where
\begin{gather}
    \mathcal{L}^{dat}=-\sum_t\sum_{k=1}^{n_{dat}}[\hat{P}_t^{dat}(dat_k) {\rm ln}{P_t^{dat}(dat_k)}] \\  
    \mathcal{L}^m(i)=-\sum_t\sum_{z \in \{0,1\}}[\hat{P}_t^{m,i}(z){\rm ln}{P_t^{m,i}(z)}] \\ 
    \mathcal{L}^v(i)=-\sum_t \sum_{k=1}^{n_{i}}[\hat{P}_t^{i}(v_k^i){\rm ln}{P_t^{i}(v_k^i)}] 
\end{gather}
where $n_{dat}$ is the number of dialogue act types,
$n_i$ is the number of values for slot $i$,
$\hat{P}_t^*$ are the gold distributions obtained from the training data, and $P_t^*$ are defined in the preceding subsection.
$\lambda$ and $\gamma$ are hyper-parameters.


%
Furthermore, we found that performance improvement can be observed when applying weak heuristic supervision on the intermediate variables, and the supervision signal can be easily obtained by simple string matching rules. This is a common practice for training sophisticated neural networks \cite{liu2016att-superversion, kiddon2016globally}. More specifically, we apply extra supervision on the update gate of the value memory (see Eq. \ref{eq:updategate}) and the attention weight of an utterance (see Eq. \ref{eq:attnweight}). 
Those intermediate supervision is applied with a two-stage training schema: firstly, we pretrain our model only with the heuristic loss ($\mathcal{L}^{att}$, see below) for several epochs, and then train the model further with the loss ($\mathcal{L}$) defined by Eq. \ref{eq:loss} for the remaining epochs.

The heuristic supervision loss is defined as follows:
\begin{align}
\label{eq:loss-att}
    \mathcal{L}^{h}=-\sum_t\sum_i\sum_{j=1}^{n_{x,t}}[\hat{\alpha}_{i,j}^t{\rm ln}{\alpha}_{i,j}^t]\notag\\
    -\sum_t\sum_i[\hat{\beta}_t^i{\rm ln}{\beta}_t^i+(1-\hat{\beta}_t^i){\rm ln}(1-{\beta}_t^i)]
\end{align}
where $n_{x,t}$ is the number of words in $x_t$ at turn $t$ and $i$ is the slot index.

Note that $\hat{\alpha}_{i,k}^t$ and $\hat{\beta}_t^i$ represent the gold distributions of the update and attention weights, respectively.
For each word $w_j$ of utterance $x_t$, if $w_j$ appears in the values of slot $i$,  $\hat{\alpha}_{i,j}^t=1$ and $\hat{\beta}_t^i=1$, otherwise
$\hat{\alpha}_{i,j}^t=0$ and
$\hat{\beta}_t^i=0$. This means that if a value of a slot appears in the utterance, the value (also the word) should be attended w.r.t. that slot, and the update weight should be equal to 1. By this way, the value memory of the corresponding slot can be updated accordingly.

\section{Experiment}
\subsection{Data Preparation}
We first evaluated our memory augmented dialogue management model on two synthetic datasets adopted from the \emph{dialog bAbI} dataset\cite{bordes2016learning} and \emph{the Second Dialogue State Tracking Challenge} dataset \cite{henderson2014second}, which are originally proposed for end-to-end dialogue systems and dialogue state tracking task. However, both of the above two datasets are small-scale. To better assess the performance of our proposed model on large-scale datasets, we collected a new Chinese dialogue management dataset consisting of real conversations from the flight booking domain.

\subsubsection{DMBD: Dialogue Management bAbI Dataset}
The original {\em dialogue bAbI} dataset (DBD) is designed to evaluate the performance of end-to-end dialogue systems on the task of restaurant reservation. In \cite{bordes2016learning}, the task is formulated as a machine comprehension task by applying the MEMN2N \cite{sukhbaatar2015end} model, considering the dialogue context and last user utterance as story and question respectively, and the system response is selected from a fixed answer set. The DBD dataset is composed of five manually constructed subtasks: {\em issuing API calls, updating API calls, displaying, providing extra information} and {\em full dialogue}, to examine the system performance on different tasks, in which the {\em full dialogue} is a combination of the first four tasks. The data for these tasks were collected through a simulator which is based on an underlying knowledge base along with some manually-crafted natural language patterns, where the simulator rules can be utilized by us to perform dialogue act annotations. For more details of DBD, please refer to \cite{bordes2016learning}.

\begin{table}[htbp]
\begin{center}
\begin{tabular}{| c | c | c |}
  \hline
    \multicolumn{2}{|c|}{\bf Informable slots} & \multicolumn{1}{c|}{\bf Requestable slots}\\
  \hline
    Name & \#Value & Name\\
  \hline
    Cuisine & 10 & Address\\
    Location & 10 & Telephone\\
    Price & 3 &\\
    Size & 4 & \\
  \hline
\end{tabular}
\end{center}
\caption{\label{tab:babi-ontology}Ontologies of the DMBD dataset. An {\em informable slot} means that user can provide values to the slot to constrain a query to KB; while a {\em requestable slot} can only be queried from KB without any user provided value.}
\end{table}

Since the dialogue act types and slot-value pairs are not annotated in DBD, we have to do this by ourselves to train our model. Fortunately, we can easily annotate the system response utterances because the original data is generated with an underlying knowledge base and some simple natural language patterns. We thus did reverse engineering by conducting automatic annotations with manually-crafted rules utilizing the knowledge base of DBD to label the dialogue act type and slot-value pairs for each utterance. This processed dataset for dialogue management is termed as {\it Dialogue Management bAbI Dataset} ({\bf DMBD}) in the following sections.

In DMBD, the original user and system utterances are reserved to serve as the input of each turn of dialogue, while the output is changed from system utterance to its dialogue act, as detailed in Table \ref{tab:dialogact}. The resulting DMBD dataset has fifteen dialogue act types, four informable slots and two requestable slots, as seen in Table \ref{tab:babi-ontology}.
An informable slot means that user can provide values to the slot to constrain a query to KB; while a requestable slot can only be queried from KB without any user provided value. Note that DMBD shares the same KB with DBD. As the requestable slots are only used for issuing API calls, in our implementation, we design a special informable slot called {\em Ask Slot}, which tracks the slots that are to be queried. The values of {\em Ask Slot} are the names of requestable slots.



\subsubsection{DM-DSTC: Dialogue Management of the Second Dialogue State Tracking Challenge dataset}
The dialogues in the above DMBD are collected via a simulator which employs hand-crafted templates, and are thus more or less synthetic. In order to evaluate the performance of our model on real-world dialogue corpus, we conducted another experiment based on DSTC2 which is a real world dialogue dataset, and it is also about the task of restaurant reservation.

The original DSTC2 dataset is for dialogue state tracking, in which the output at each turn is the filled slots and their values which have already been presented by the user so far. The dialogue act of the system utterance is also annotated and is thus directly utilized as model output.
We thus transform the original DSTC2 dataset to our settings for dialogue management, referred to as {\bf DM-DSTC}. The ontologies of dialogue act type and slot in the original dataset are directly reused in the DN-DSTC.

The resulting DM-DSTC is composed of four informable and nine requestable slots, and the average value number of informable slots is 54, which is much higher than that of DMBD, and the enhanced complexity of DM-DSTC dataset reflects the characteristics of real-world data which is more stochastic and noisy. We also created a special slot for requestable slots in this experiment as we did in the DMBD experiment. Some statistics of DM-DSTC are shown in Table \ref{tab:dstc-ontology}.

\begin{table}[htbp]
\begin{center}
\begin{tabular}{| c | c | c |}
  \hline
    \multicolumn{2}{|c|}{\bf Informable slots} & \multicolumn{1}{c|}{\bf Requestable slots}\\
  \hline
    Name & \#Value & Name\\
  \hline
    Food & 91 & Addr, Area, Food\\
    Pricerange & 3 & Phone, Pricerange\\
    Res\_name & 113 & Postcode, Signature\\
    Area & 5 & Res\_name\\
  \hline
\end{tabular}
\caption{\label{tab:dstc-ontology}Ontology of the DM-DSTC dataset. The {\em Res\_name} indicates restaurant name. The average value number of informable slots is 54 which is much higher than that of DMBD dataset. The enhanced complexity of DM-DSTC reflects the characteristics of real-world dialogue data.}
\end{center}
\end{table}

\subsubsection{ALDM: Alibaba Dialogue Management Dataset}
\label{sec:aldm-dataset}
The sizes of the above two datasets are limited, we thus propose ALDM to test our model's performance on large-scale dataset.
ALDM is a Chinese dataset, consisting of real conversations from the flight-booking domain, in which the system is supposed to acquire departure city, arrive city and departure date information from the user to book a flight ticket. To better fit our model, the departure date values in the corpus are preprocessed into an uniform MM.DD format, e.g., {\em 12.25} for {\em 25th, Dec.}. ALDM is much larger than the other two datasets, where there are 15,330 sessions for training, 7,665 for validation, and 3,832 for test. 
On average, there are 5 turns in a session. The average sentence length is 4, and particularly, most of the user responses have only one word as users only provide the departure or arrival city, or the departure data.
One difference to the other two datasets exists in that the departure city slot and the arrive city slot share the same value list, which raises additional difficulty to require the model to identify which slot the city name in the user utterance should be filled in. To handle this issue, the model should be able to fill slots conditioned on the dialogue context. For example, if the user responds with {\em Beijing} to the last system response {\em Where are you flying from?}, the value of {\em Beijing} should be filled in the {\em departure city}. Another difference is that there are not requestable slots due to the fact that ALDM is system-driven.

\begin{table}[htbp]
\begin{center}
\begin{tabular}{| c | c | c |}
  \hline
    {\bf DA type} & \multicolumn{2}{c|}{\bf Informable Slots}\\
  \hline
    ask\_dep\_loc & Name & \#Value\\
  \cline{2-3}
    ask\_arr\_loc & Dep\_city & 174 \\
    ask\_dep\_date & Arr\_city & 174 \\
    offer, end & Date & 100 \\
  \hline
\end{tabular}
\caption{\label{tab:aldm-ontology}Ontology of the ALDM dataset. The {\em ask\_} DA type means the system is asking the user for information, {\em offer} means the system is giving recommendation and {\em end} means the dialogue session is done. {\em Dep\_city} and {\em Arr\_city} represent the slot of departure city and arrive city respectively, and they share the same value list. The value of {\em Date} slot is transformed into a uniform MM.DD format.}
\end{center}
\end{table}

As shown in Table \ref{tab:aldm-ontology}, ALDM is composed of 3 informable slots, and the average value number is 150, which is remarkably larger than those of the above two datasets. And there are 5 dialogue act types as shown in Table \ref{tab:aldm-ontology}.

\subsection{Experimental Setup}
Our model is implemented with Tensorflow \cite{abadi2016tensorflow}. The word embeddings used in each dataset were pretrained on their own dialogue corpora, where there are 15,000 sessions in DMBD (3,000 per each task), 2,118 sessions in DM-DSTC and 26,827 sessions in ALDM, using the GloVe algorithm \cite{pennington2014glove}. The dimensions of word embeddings, memory column vectors, and state vectors were all set to 128, and there are 8 columns in the external memory. 
We first pretrain our model with the heuristic loss $\mathcal{L}^h$ (see Eq. \ref{eq:loss-att}) for 2 epochs and then continue to train it using $\mathcal{L}$ in Eq. \ref{eq:loss}.

The parameters $\gamma$ amd $\lambda$ in $\mathcal{L}$ are not constant during training.
More specifically, in the first 7 epochs, $\lambda$ increases linearly from 0 to 1 while $\gamma$ remains zero, and in the following 7 epoches $\gamma$ also rises from 0 to 1 linearly with $\lambda$ unchanged. The reason for this setting is that the process of the value update in the slot-value memory has strong influence on the training of other components.
All the other parameters are initialized with a random uniform distribution $\mathcal{N}(0, 1)$.

We used the train/valid/test partition of the original DBD for each task, where there are 1,000 sessions in each set; and the partition of DM-DSTC is 1412/353/353. For ALDM, we split the dataset into 15,330/7,665/3,832.

We trained our model using ADAM \cite{kingma2014adam} with a learning rate which is set to 0.002, and the momentum parameters ${\beta}_1 = 0.9$ and ${\beta}_2 = 0.999$. For each dataset, the model is trained with at most 15 epochs. We use the model parameter with the lowest validation loss for test.

\subsection{Baseline}

We included two types of baselines in the evaluation. The first type is to select a sentence as answer from a predefined candidate answer set in a machine comprehension manner, as described in \cite{bordes2016learning}. The second type is to predict a structured dialogue act, the same as our model, where the models need to make predictions over all combinations of dialogue act type and slot-value pairs.

In the baselines of the first type, each candidate answer sentence is a natural language utterance, which lexicalizes\footnote{Lexicalizing a dialogue act means converting the act from formal semantic representation to a natural language utterance.} an underlying dialogue act. However, the candidate answer set is not complete, where not all possible combinations of dialogue act type and slot-value pairs are included. In other words, the size of the answer space in the first type is less than that in the second type. Thus, the first setting is therefore easier than the second one.

The baselines of the first type, which select an utterance from a predefined candidate answer set \cite{bordes2016learning}, are listed as follows:
\begin{itemize}

\item {\bf TF-IDF}: A TF-IDF matching algorithm\cite{salton1986introduction} which computes a cosine similarity score between the input (the whole dialogue history) and a candidate sentence, and the sentence with the highest score is selected as the final answer. Both the input and the candidate sentence are represented by the average of bag-of-word vectors.
    
\item {\bf TF-IDF(+ type)}: An enhanced version of TF-IDF by introducing additional entity type features.

\item {\bf Supervised Ebd}: An information retrieval model based on trainable word embeddings. The similarity score between an input and a candidate sentence is the inner product of their averaged word embeddings. The  is trained with a margin ranking loss \cite{bai2009supervised}.

\item {\bf MEMN2N}: Standard end-to-end memory networks \cite{sukhbaatar2015end, bordes2016learning}. It stores the dialogue history information in a memory network and chooses a response by running multi-hop reasoning upon the history. 

\item {\bf MEMN2N(+ match)}: A variant of MEMN2N which included additional features about entity types.
\end{itemize}

The baselines of the second type, which predict a structured dialogue act, the same as our proposed model, are as follows:

\begin{itemize}

\item {\bf MEM}: A memory network model which predicts dialogue act. For each output structure (DA type, slot-value, and mask), a MEMN2N is introduced to make prediction.

\item {\bf RNN}: A recurrent neural network model with turn-level input and output. The dialogue act predictions (type and slot-value) are based on the hidden state $\bm{S}_t$ at each time step $t$.

\item {\bf MAD - SM}: A variant of our proposed model without the slot-value memory. Those predictions involving the slot-value memory are modified to using only the memory controller state $\bm{S}_t$ to make prediction.

\item {\bf MAD - Attn}: A variant of our model without the slot-level attention mechanism. In this setting, the averaged word embeddings of an utterance is used to update the slot-value memory.

\item {\bf MAD - EM}: A variant of our model without the external memory. The predictions involving the external memory are modified to using the memory controller state $\bm{S}_t$ only, just as MAD-SM.
\\
\end{itemize}

It should be noted that the MEMN2N and MEM baseline take as input a context-question pair at each round, which means they have to make calculation on the cumulated dialogue context at each turn. Thus with the increasing of the dialogue context, there is an exponential increase in the computation complexity. While for our model, the context information is stored in the memory network, and the computation time in each turn is basically the same.

\subsection{Performance on DMBD}

In this section, we evaluated the performance of our model and the baselines on the DMBD dataset. The prediction accuracy on both turn-level and session-level evaluation is reported, similar to \cite{bordes2016learning}. Based on the distribution defined in Section \ref{sec:prediction}, our model chooses a dialogue act with the maximal probability as output, respectively for {\em DA type}, {\em slot-value} and {\em mask}. Note here that for  DA type and mask, the prediction is judged as correct only if the output matches the target. 
As mentioned in Section \ref{sec:taskdef}, mask is an auxiliary variable helping to filter the undesired slot-value pairs in a predicted dialogue act.
Thus for the prediction of slot-value, we only need to correctly predict those slot-value pairs whose mask value is 1.
Finally, the overall dialogue act is correct only if its DA type, slot-value and mask are all correctly predicted. And a dialogue session is correct only if all the dialogue acts in the session are correctly predicted. We termed this session-level evaluation. 

\begin{table*}
\centering
\begin{tabular}{c | c c | c c | c c | c c | c c }
    \hline
        \multirow{2}{*}{Metrics} & \multicolumn{2}{c|}{1 Issuing } & \multicolumn{2}{c|}{2 Updating} & \multicolumn{2}{c|}{3 Displaying} & \multicolumn{2}{c|}{4 Providing} & \multicolumn{2}{c}{5 Full} \\ 
        & \multicolumn{2}{c|}{API calls} & \multicolumn{2}{c|}{API calls} & \multicolumn{2}{c|}{options} & \multicolumn{2}{c|}{options} & \multicolumn{2}{c}{dialogs} \\
    \hline
        {TF-IDF (no type)} & 5.6 & (0) & 3.4 & (0) & 8.0 & (0) & 9.5 & (0) & 4.6 & (0)\\
        {TF-IDF (+ type)} & 22.4 & (0) & 16.4 & (0) & 8.0 & (0) & 17.8 & (0) & 8.1 & (0)\\
        {Nearest Neighbor} & 55.1 & (0) & 68.3 & (0) & 58.8 & (0) & 28.6 & (0) & 57.1 & (0)\\
        {Supervised Ebd} & 100 & (100) & 68.4 & (0) & 64.9 & (0) & 57.2 & (0) & 75.4 & (0)\\
        {MEMN2N (no match)} & 99.9 & (99.6) & 100 & (100) & 74.9 & (2.0) & 59.5 & (3.0) & 96.1 & (49.4)\\
        {MEMN2N (+ match)} & 100 & (100) & 98.3 & (83.9) & 74.9 & (0.0) & 100 & (100) & 93.4 & (19.7)\\
    \hline
        {MEM} & 47.4 & (0.1) & 61.1 & (0.1) & 24.6 & (0.1) & 56.7 & (0.8) & 25.2 & (0.1) \\
        {RNN} & 80.6 & (0.1) & 45.5 & (0.0) & 30.0 & (0.0) & 57.2 & (0.0) & 3.7 & (0.0) \\
    \hline
        {MAD} & 99.0 & (94.2) & 100 & (100) & {\bf 99.1} & {\bf (90.6)} & 100 & (100) & {\bf 99.9} & {\bf (97.8)}\\
    \hline
\end{tabular}
\caption{\label{tab:result} The accuracy across all tasks and methods. The numbers in brackets are the accuracy at the session level, and numbers without brackets are at the turn level. A session is correct only if all the sentences in the session are predicted correctly.}
\end{table*}

\subsubsection{Overall Performance Analysis}

We first evaluated our proposed model based on the overall accuracy of dialogue act prediction, as shown in Table \ref{tab:result}. The results of baselines of the first type are reprinted from their original paper \cite{bordes2016learning}, because the partitions of training/validation/test data are the same as ours, and the results are hence directly comparable. Both turn-level and session-level results on all the five tasks are reported. 
We have the following observations:
\begin{itemize}
    \item 
    MAD obtains the best performance on most of the tasks. The model obtains an accuracy of about 100\% at both turn and session-level evaluation, which shows the effectiveness of our proposed model. While in Task 1, MAD is at the second place, where Supervised Ebd and MEMN2N (+match) methods obtains 100\% accuracy at both turn and session-level evaluation, which is 1\% higher than ours. MAD's defect on Task 1 can be attributed to a potential rule in Task 1, where if the user doesn't provide enough values to form a query, the agent will request for the value of slots in a fixed order. For example in task 1, the agent requests for slots in an order of ({\em Cuisine} $\rightarrow$ {\em Location} $\rightarrow$ {\em Size} $\rightarrow$ {\em Price}). However, this order rule is not essential for a practica application, where the agent can request for values in an arbitrary order as long as it can obtain all necessary values.

    \item

\end{itemize}

\subsubsection{Fine-grained Performance Analysis}

To better understand how the slot-value memory and the external memory influence the performance, we further analyzed the fine-grained prediction accuracy of MAD and its variants in addition to the overall dialogue act prediction. Evaluation on the fine-grained predictions is shown in Table \ref{tab:subtask}. We have the following observations:


\begin{itemize}
    \item The variants of MAD, MAD-SM, which ablates the slot-value memory module, obtains degraded performance on overall accuracy compared to MAD. MAD-Attn, which removes the slot-level attention mechanism, works worse than MAD but still slightly better than MAD-SM on each task. The performance of MAD-EM drops even more than MAD-SM on all tasks except for Task 1. The RNN model, which can be regarded as MAD without slot-memory and external memory, performs even worse on most of the 5 tasks.

    \item The fine-grained results demonstrate the effectiveness of our proposed model more specifically. Here we can see that the accuracy of MAD on both slot-value and mask is 100\%, while the prediction on DA type has very few errors. The high accuracy of slot-value prediction indicates that the slot addressing and the attentive question representation work well, which is attributed to the slot-value memory and attention supervision we applied. The contribution of the external memory is also shown by the high performance of DA type and mask prediction.
    
    \item The slot-value memory leads to significant improvements in slot-value accuracy. In our model, the role of the slot-value memory is to extract semantic information about slots during the dialogue, thus the ability of tracking slot-value information should decrease if the slot-value memory is removed. As shown in Table \ref{tab:subtask}, the prediction accuracy of MAD-SM on slot-value drops much from 100\% to around 30\%. However, the performance on dialogue act type and mask prediction are not heavily affected, and the accuracy is still above 90\%.
    
    \item The slot-level attention mechanism we applied on semantic information extraction influences the performance remarkably. In MAD-Attn, the slot-level attention mechanism is removed, and the value update is based on averaged word embeddings of user utterance. Intuitively, the update of the slot-value memory is not able to concentrate on relevant words without attention mechanism, thus the performance of slot-value prediction must be heavily influenced. The experiment results also support our hypothesis, where the accuracy of slot-value prediction degrades remarkably, but is still better than that of MAD-SM since MAD-Attn retains the slot-value memory. The attention mechanism affects dialogue act type and mask prediction very slightly.
    
    \item The external memory significantly improves the performance of DA type and mask accuracy by enhancing the representation capacity of the original RNN state. In MAD-EM, the external memory is removed, and those predictions involving the external memory, that is the prediction of DA type and mask, are changed to use the memory controller state, which is identical to the hidden state in a RNN model. Compared to MAD, the accuracy of MAD-EM on DA type and slot-value prediction decreases heavily. This is attributed to the enhanced representation capacity, meaning that the model can do better in capturing longer term temporal dependencies in dialogue.
\end{itemize}

From the above analysis, we can see that the effect of the slot-value memory is mainly on predicting slot-value, while the effect of the external memory is on predicting dialogue act type and mask.
However, the influence of the modules on the performance is more complex. We can see from Table \ref{tab:subtask} that DA type and mask accuracy will also decrease if the slot-value memory is removed, and so will slot-value accuracy when we remove external memory. This means the two memory networks in our model are coupled correlatively by the memory controller and can affect the performance of each other.

%

\begin{table}[htbp]
\small
\begin{center}
\begin{tabular}{ c | c | c c | c c | c c | c c | c c }
  \hline
    \multicolumn{2}{c|}{\bf Task} & \multicolumn{2}{c|}{1} & \multicolumn{2}{c|}{2} & \multicolumn{2}{c|}{3} & \multicolumn{2}{c|}{4} & \multicolumn{2}{c}{5}\\
  \hline
  \multirow{4}{*}{\bf DA type} & MAD-SM & 93.9 & (65.9) & 100 & (100) & 95.6 & (58.2) & 100 & (100) & 90.9 & (11.9) \\
    & MAD-EM & 95.8 & (80.5) & 65.7 & (3.5) & 56.3 & (5.8) & 100 & (100) & 17.8 & (0) \\
    & MAD-Attn & {\bf 99.5} & (96.9) & 100 & (100) & 99.0 & (90.3) & 100 & (100) & {\bf 99.9} & (98.6) \\
    & MAD & 99.0 & (94.2) & 100 & (100) & {\bf 99.1} & (90.6) & 100 & (100) & 99.9 & (97.8) \\
  \hline
  \multirow{4}{*}{\bf slot-value} & MAD-SM & 21.1 & (0.3) & 22.3 & (0) & 18.4 & (0) & 40.3 & (0.1) & 20.9 & (0) \\
    & MAD-EM & 100 & (100) & 95.3 & (65.8) & 27.5 & (0.1) & 100 & (100) & 22.6 & (0) \\
    & MAD-Attn & 26.8 & (0.5) & 24.8 & (0) & 27.5 & (0) & 41.3 & (0.1) & 31.4 & (0) \\
    & MAD & 100 & (100) & 100 & (100) & 100 & (100) & 100 & (100) & 100 & (100) \\
  \hline
  \multirow{4}{*}{\bf mask} & MAD-SM & 1.0 & (1.0) & 100 & (100) & 99.9 & (99.9) & 100 & (100) & 98.8 & (6) \\
    & MAD-EM & 99.1 & (88.8) & 87.8 & (2.6) & 87.8 & (16.4) & 100 & (100) & 66.8 & (0) \\
    & MAD-Attn & 100 & (100) & 100 & (100) & 100 & (100) & 100 & (100) & 100 & (100) \\
    & MAD & 100 & (100) & 100 & (100) & 100 & (100) & 100 & (100) & 100 & (100) \\
  \hline
  \multirow{4}{*}{\bf Overall} & MAD-SM & 77.2 & (0.2) & 78.9 & (0) & 70.7 & (0) & 57.3 & (0.1) & 59.6 & (0) \\
    & MAD-EM & 95.2 & (78.2) & 57.4 & (0.2) & 40.5 & (0.0) & 1.0 & (1.0) & 3.1 & (0.0)\\
    & MAD-Attn & 82.7 & (0.5) & 79.0 & (0) & 73.9 & (0) & 57.3 & (0.1) & 67.7 & (0) \\
    & MAD & {\bf 99.0} & (94.2) & 100 & (100) & {\bf 99.1} & (90.6) & 100 & (100) & {\bf 99.9} & (97.8)\\
  \hline
\end{tabular}

\end{center}
\caption{\label{tab:subtask}Fine-grained performance on the DMBD dataset. We tested the performance of our proposed model and three of its variations on both turn and session level, where for each model the dialogue act type, slot-value, mask and overall prediction accuracy on each task is reported. The highest accuracy on turn level which is lower than 100\% is in bold font.}
\end{table}

\subsection{Performance on DM-DSTC}
\label{sec:per-dm-dstc}
Although our proposed model obtains good results on DMBD, it should be noted that the performance reflected by the above results are somehow optimistic due to two facts: First, these dialogues are generated by rules, which are much simpler than real dialogue data. Second, the number of slots and values in DMBD is quite small, while in real applications the number may become very large.

To assess the performance of our proposed model on real dialogue data, we conducted another experiment on DM-DSTC. Different from DMBD, there is only one task in the DM-DSTC dataset. We only reported the results of the methods which predict dialogue act as output. It should be pointed out that in this new dataset, many values in dialogue act annotation didn't appear exactly in user utterances (such as {\em asian oriental}), thus for those values we can not provide precise attention supervision, which will affect the performance of slot-level attention. 
Moreover, the {\em Res\_name} slot in this dataset degrades the accuracy because its value does not appear in the dialogue context, and is queried from a knowledge base conditioned on previous search constraints, which is not consistent with our model setting. We reported the fine-grained and overall accuracy at the turn level and session level, as shown in Table \ref{tab:dstc}.


\begin{table}[htbp]
\small
\begin{center}
\begin{tabular}{ c | c c | c c | c c | c c}
  \hline
    Metrics & \multicolumn{2}{c|}{DA type} & \multicolumn{2}{c|}{slot-value} & \multicolumn{2}{c|}{mask} & \multicolumn{2}{c}{All}\\
  \hline
    MEM & 62.5 & (9.9) & 14.2 & (0.0) & 71.0 & (0.1) & 0 & (0.0)\\   
    RNN & 50.9 & (0.3) & 14.3 & (0.1) & 61.8 & (0.3) & 0.1 & (0.0) \\
    MAD-SM & 64.1 & (13.6) & 11.6 & (0.1) & 81.6 & (0.4) & 17.1 & (0.1)\\
    MAD-Attn & {\bf 64.6} & (12.5) & 18.5 & (0.1) & 80.8 & (1.0) & 16.9 & (0.0)\\
    MAD-EM & 44.9 & (2.3) & 17.5 & (0.1) & 69.7 & (0) & 5.7 & (0.0)\\
  \hline
    {\bf MAD} & 63.8 & (11.0) & {\bf 27.3} & (0.1) & {\bf 82.1} & (1.3) & {\bf 18.8} & (0)\\ 
  \hline
\end{tabular}
\end{center}
\caption{\label{tab:dstc}Fine-grained and overall accuracy on the DM-DSTC dataset. The number in brackets are the accuracy at the session level, and number without brackets are at the turn level.}
\end{table}

The results in Table \ref{tab:dstc} demonstrate our model is still comparable to the vanilla memory network model. Compared to MEM and RNN, our proposed method obtains higher accuracy on turn-level overall prediction, as well as the dialogue act type and mask prediction.
Although MEM's accuracy on DA type , slot-value and mask prediction is slightly lower than ours, its overall accuracy on turn-level is far less than our proposed model. This can be attributed to the framework of MEM, where its DA type, mask and slot-value prediction is trained separately, while in our model these three tasks are trained.
For the variants of MAD, the experiment results are consistent with what we observed in DMBD. MAD-SM obtains lower accuracy on slot-value prediction compared to MAD, while maintains similar accuracy on DA type and mask. For MAD-Attn, the result is similar to MAD-SM when compared to MAD, but its accuracy on slot-value prediction is obviously higher than that of MAD-SM since it maintains the slot-value memory network. MAD-EM, which removes the external memory, obtains significantly lower accuracy on the prediction of dialogue act type and mask, and its accuracy on slot-value prediction is also reduced.

We can see that the performance of slot-value prediction is the bottleneck of promoting overall accuracy. That can be attributed to the data feature of DM-DSTC, where many values of slots does not appear precisely in the user utterance, which makes it hard to acquire accurate attention supervision, thus the model's capacity of extracting semantic features from user utterance is negatively influenced. For the prediction of DA type and mask, although the result is far better than that of slot-value, the accuracy is still not so high as that in DMBD. This can be attributed to the characteristics of real-world data, where there exists much more probability uncertainty and noise than DMBD. More specifically, in different sessions, the DA type of agent response varies much even it is given the same dialogue context. What's more, the agent response in original DSTC2 dataset is conditioned on the knowledge base query result which is not provided, and this also restricts our model's ability on predicting DA type and mask.
\subsection{Performance on ALDM}
\begin{table}[htbp]
\small
\begin{center}
\begin{tabular}{ c | c c | c c | c c | c c}
  \hline
    Metrics & \multicolumn{2}{c|}{DA type} & \multicolumn{2}{c|}{Slot-value} & \multicolumn{2}{c|}{Mask} & \multicolumn{2}{c}{All}\\
  \hline
    MEM & 64.9 & (1.4) & 73.5 & (0.0) & 100.0 & (100.0) & 0.0 & (0.0)\\
    RNN & 60.0 & (0.0) & 80.0 & (0.0) & 100.0 & (100.0) & 40.0 & (0.0)\\
    MAD-SM & 60.3 & (0.0) & 80.0 & (0.0) & 100.0 & (100.0) & 40.3 & (0.0)\\
    MAD-Attn & 76.4 & (15.7) & 100.0 & (100.0) & 100.0 & (100.0) & 76.4 & (17.1)\\
    MAD-EM & 76.4 & (15.4) & 98.6 & (92.8) & 100.0 & (100.0) & 74.9 & (14.2)\\
  \hline
    {\bf MAD} & {\bf 76.7} & (16.3) & 100.0 & (100.0) & 100.0 & (100.0) & {\bf 76.7} & (16.3)\\
  \hline
\end{tabular}
\end{center}
\caption{\label{tab:aldm}Fine-grained and overall accuracy on the ALDM dataset. The number in 
bracket is the accuracy at the session level, and the number without bracket is at the turn level.}
\end{table}

We reported the results of the methods which can output a structured dialogue act as we did in Section \ref{sec:per-dm-dstc}. The mask prediction is relatively simple for ALDM in which most of the slot values only appear in the last system response, and thus all the models have an accuracy of 100\%. Therefore, the following analysis will be focused on the DA type and slot-value.

A difference of ALDM compared to the other two datasets is that ALDM is more system-driven, which makes it hard for our model to correctly predict the order of {\em ask\_} DA type, 
For instance, {\em ask\_dep\_Loc} is only based on the currently filled slots. If the departure location is provided by the user, the system can ask for either the arrive location or the departure date in the next turn, which makes the next DA type difficult to predict. Thus the DA type accuracy is not as good as that in DMBD. However, when $N-1$ slots is already filled (N is the total number of slots to complete a booking task), the next slot to be asked is determinate. Thus, the dialogue state still has impact on DA type prediction, which is shown by the results of MAD-SM and RNN in which the two models removed the slot-value memory.

Although the average number of the slot values in ALDM is much larger than that in the other two dataset, we still obtain high slot-value accuracy. This can be attributed to the high data quality of ALDM which is carefully cleaned before training. By removing the slot-value memory (RNN and MAD-SM) we can see that the slot-value accuracy decreases remarkably, which shows the ability of slot-value memory for maintaining dialogue states. As it can be seen from Table \ref{tab:aldm}, the slot-value accuracy of our full model is the same as that of {\em MAD-Attn}. This is because of the nature of the ALDM dataset that the user responses are mainly one-word sentences, which makes no difference between the models with/without attention mechanism.

\begin{table}[htbp]
\small
\begin{center}
\begin{tabular}{ c | c c }
  \hline
    Metrics & {Departure-City} & {Arrive-City}\\
  \hline
    MEM & 2.7 & 4.1  \\ 
    RNN & 0.2  & 0.1  \\
    MAD-SM & 0.5 & 0.3 \\
    MAD-Attn & 100.0  & 100.0  \\
    MAD-EM & 96.5  & 96.2  \\
  \hline
    {\bf MAD} & 100.0  & 100.0 \\
  \hline
\end{tabular}
\end{center}
\caption{\label{tab:departure-arrive}Prediction accuracy on the departure city and arrive city slots. The number in bracket is the accuracy at the session level, and that without bracket at the turn level.}
\end{table}

To verify the model's ability to combine context information in slot filling, we further analyzed the prediction accuracy on the {\em Departure\_City} slot and the {\em Arrive\_City} slot. As described in Section \ref{sec:aldm-dataset}, they share the same value list. The ability of identifying values from different slots is mainly controlled by the update gate ${\beta}_t^i$ as defined in Section \ref{sec:sv-memory}. Slot-value memory dominates the prediction of the next slot values, which can be seen from the results of MAD-SM, RNN, and MEM in Table  \ref{tab:departure-arrive}. The results drop dramatically when removing the slot-value memory (RNN and MAD-EM).
For MEM, although its accuracy is higher than that of RNN and MAD-EM, it's still much lower than our proposed model. This is because that 1) the city number is too large for MEM to predict, and 2) MEM fails to identify which slot the value belongs to.

\subsection{Parameter Tuning}

Generally speaking, the performance of neural network models is highly correlated with the number of parameters. There are many important hyper-parameters in our model, including the dimensions of the slot-value memory and external memory, and the number of column vectors in the external memory. We evaluated the influence of these hyper-parameters on performance. The following experiments were performed on the DM-DSTC dataset.

\begin{figure}
  \includegraphics[width=0.6\textwidth]{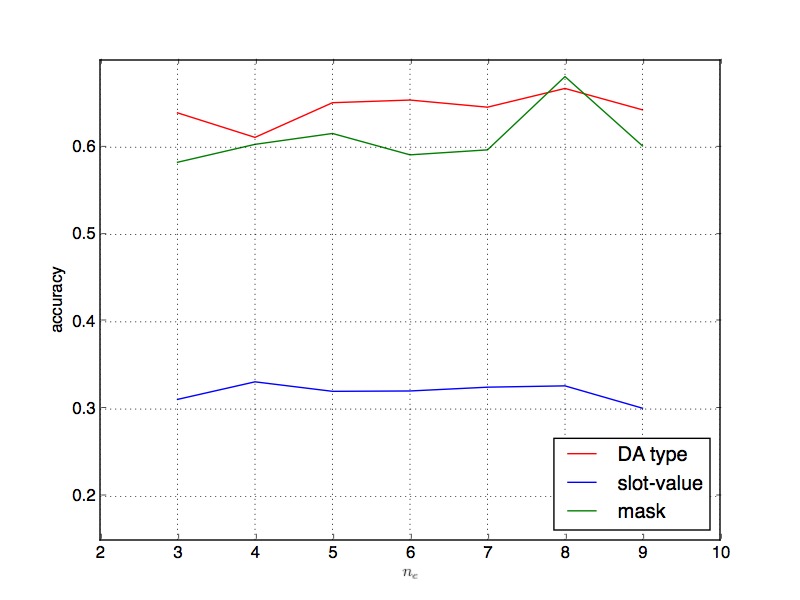}
  \caption{Fine-grained prediction accuracy on DMBD with different $n_e$ (the number of column vectors in the external memory). The optimal number is 8.
  }
  \label{fig:n_e}
\end{figure}

First, we studied how the performance is influenced by the number of column vectors in the external memory $n_e$. The number $n_e$ varies from 3 to 9, with a step size of 1. We studied the accuracy change on dialogue act type, slot-value, and mask, as shown in Figure \ref{fig:n_e}. For predicting dialogue act type and mask, the optimal $n_e$ is 8 and the optimal accuracy is significantly better than others. For predicting slot-values, although the optimal $n_e$ is 4 with an accuracy of 0.331, the accuracy is almost the same (from 0.321 to 0.331) when varying $n_e$ from 4 to 8.

\begin{figure}
  \includegraphics[width=0.6\textwidth]{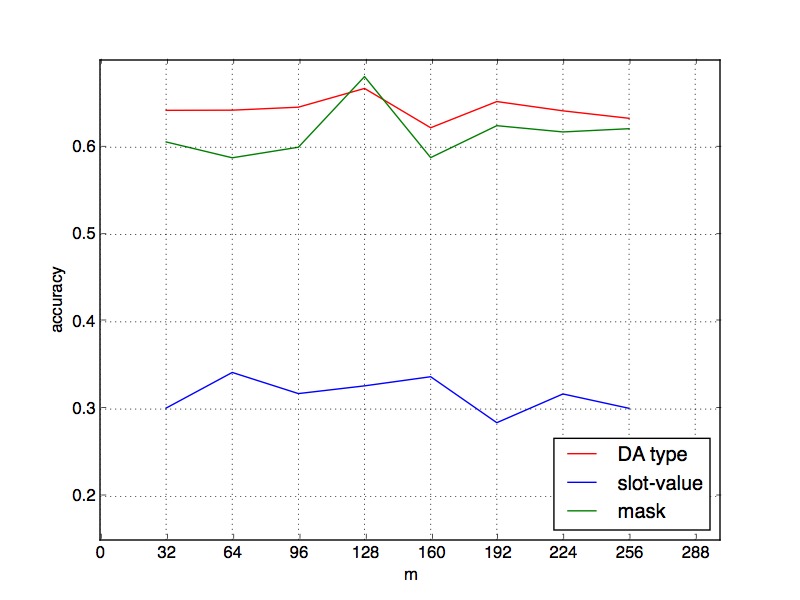}
  \caption{Accuracy change on DMBD with different dimensions of the column vectors in the external memory. The optimal number is 128.
  }
  \label{fig:dimension}
\end{figure}

Second, we studied the influence of the dimension of column vectors, as shown in Figure. \ref{fig:dimension}. The dimension number in our experiment ranges from 32 to 256 with a step size of 32. The accuracy of dialogue act type and mask is highly correlated, whose best accuracy are both obtained with the dimension of 128. While the optimal value for slot-value accuracy is obtained with the dimension of 64.



\subsection{Visualization Analysis}
\begin{figure}
\centering
\includegraphics[width=\textwidth]{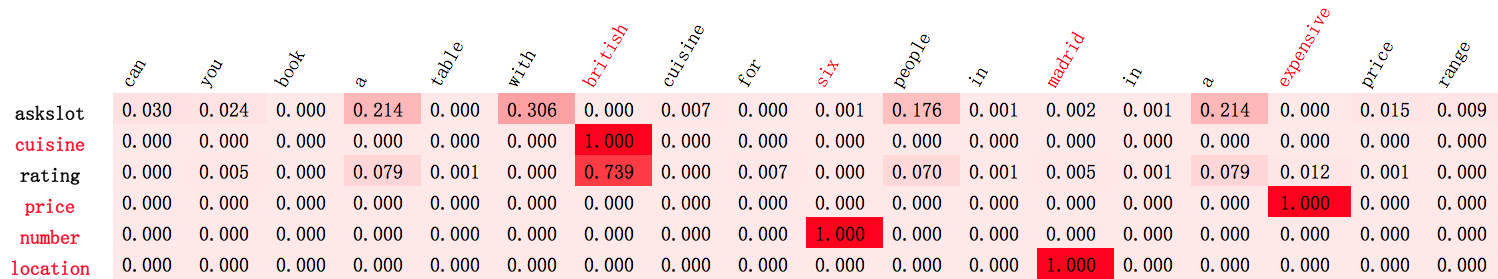}
\caption{Attention visualization. For each slot, the attention weights (in a row) are a distribution over the words of an utterance. For utterance {\em "can you book a table with british cuisine for six people in madrid in an expensive price range."} the predicted slot-value pairs are $<cuisine, british>$, $<number, six>$, $<location,madrid>$, and $<price, expensive>$. }
\label{fig:attention}
\end{figure}

Figure \ref{fig:attention} illustrates an example of the slot-level attention mechanism. 
For each slot, the model generates a distribution over the words of an utterance. Each row is thus a probability distribution over words, where the largest probability corresponds to the word that should be attended mostly. 
For utterance {\em "can you book a table with British cuisine for six people in Madrid in an expensive price range"}, 
for slot {\em Cuisine}, the most attended word is {\em British}, while for slot {\em Price }, the word is {\em expensive}, and for slot {\em Number}, the word is {\em six}. 
Note that the weight of {\em $<Rating,british>$} is also large, which is wrong intuitively in that Rating information has not yet been mentioned. However, this kind of wrong attention weight does not have influence on model performance. 
In other words, the inclusion of a slot-value pair in the predicted dialogue act is decided by two distributions: the value distribution and the slot mask distribution for a slot, as mentioned in Section \ref{sec:prediction}. The effect of faulty attention will be filtered out by mask when deciding which slots are to be addressed in final dialogue act. 

\begin{figure}
  \small
  \renewcommand\floatpagefraction{.9}   
\renewcommand\topfraction{.9}  
\renewcommand\bottomfraction{.9}  
\renewcommand\textfraction{.1}
  \includegraphics[width=0.8\textwidth]{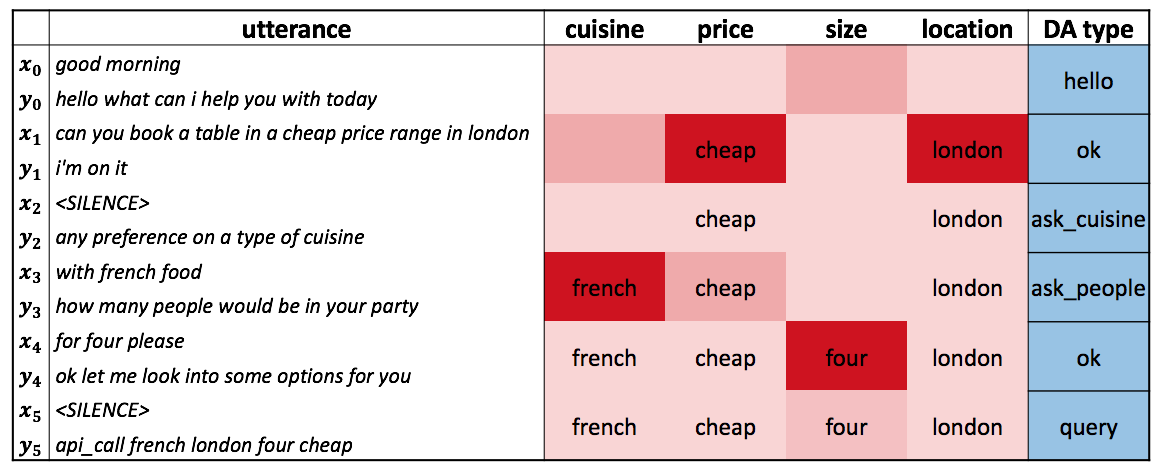}
  \caption{An example of DA prediction for a dialogue session. $\bm{x}$ represents user utterance and ${y}$ system response.The values of slots at each turn are predicted by Eq. \ref{predict:sv}. The color darkness of each cell represents the value of ${\beta}_t^i$ defined in Eq. \ref{eq:updategate}. Darker colors indicate larger values.}
  \label{fig:full-dialog}
\end{figure}

Figure \ref{fig:full-dialog} illustrates the change of the dialogue state and the predicted next dialogue act in an exemplar dialogue session. We visualized the values stored in the slot-value memory and shown the next dialogue act type predicted by the model.
At each turn, the model computes an update gate $\beta_t^i$ (Eq.\ref{eq:updategate}) for each slot $i$.
If a certain value of slot $i$ appears in user utterance $x_t$, $\beta_t^i$ increases, and the color of the corresponding cell becomes darker.
The darkness of a cell represents the value of $\beta_t^i \in [0, 1]$, which is calculated independently for each slot $i$ at each turn $t$.
The value in each cell is computed by Eq. \ref{predict:sv} and we only output the value for slot $i$ if $\beta_{\tau}^i > 0.5$ for some turn $\tau$. 
These values compose a search constraint at each turn.
In the exemplar dialogue session, each value in user utterance is captured by the attention mechanism of a user utterance, 
and its values are filled into $M^V$ with large $\beta_t^i$s.

For instance, when the user asks {\em can you book a table in a cheap price range in london?}, the {\em price} slot is filled with the value of {\em cheap }, and the {\em location} slot is filled with the value of {\em london}. The model predicts the next dialogue act {\em ask\_cuisine} which prompts the user on the preference of {\em cuisine}. As the user supplied new information with the utterance {\em with french food}, the {\em cuisine} slot is filled with the value of {\em french}. At this state, the model predicts the next dialogue act  {\em ask\_people} which should ask the user about how many people are involved. As the dialogue proceeds, the slot-value memory explicitly tracks the dialogue state, and the next dialogue act is also predicted according to the state. 

\section{Conclusion}
In this paper, we present a memory augmented dialogue management model for capturing long-range dialogue semantics by explicitly memorizing and updating the dialogue act types and slot-value pairs during interactions in task-oriented dialogue systems. 
The model employs two memory modules, namely the slot-value memory and external memory, to address the history semantics during the entire dialogue session. The slot-value memory tracks the dialogue state by memorizing and updating the values of semantic slots, and the external memory augments the single state representation of RNN by storing more context information.
We also propose a slot-level attention mechanism for attentive read of a user utterance to update the slot-value memory. The attention mechanism helps to extract the slot-related information that is addressed in a user utterance. 
Through the attention mechanism and the memory modules, our proposed model can better interpret the dialogue context in a more observable and explainable way, which also helps to predict the next dialogue act given the current dialogue state.
Results show that our model is better than the state-of-the-art baselines, and moreover, the model can offer more observable dialogue semantics by presenting
predicted slot-value pairs at each dialogue turn.
We believe that research on interactive IR may benefit from our work, particularly from the idea of enhancing the interpretability of dialogue management.


\bibliographystyle{ACM-Reference-Format}
\bibliography{sample-bibliography}